\documentclass[final]{cvpr}

\usepackage{times}
\usepackage{epsfig}
\usepackage{graphicx}
\usepackage{amsmath}
\usepackage{amssymb}
\usepackage[font=small]{caption}
\usepackage{subfigure}
\usepackage{multirow}
\usepackage[hang,flushmargin]{footmisc}
\usepackage[pagebackref=true,breaklinks=true,letterpaper=true,colorlinks,bookmarks=false]{hyperref}

\DeclareMathOperator*{\argmin}{argmin}

\begin{document}

\title{GLEAN: Generative Latent Bank for Large-Factor Image Super-Resolution}
\author{Kelvin C.K. Chan$^{1}$
\quad
Xintao Wang$^{2}$
\quad
Xiangyu Xu$^{1}$
\quad
Jinwei Gu$^{3}$
\quad
Chen Change Loy$^{1*}$\\
$^{1}$Nanyang Technological University, Singapore\\
$^{2}$Applied Research Center, Tencent PCG\quad
$^{3}$SenseBrain\\
{\tt\small \{chan0899, xiangyu.xu, ccloy\}@ntu.edu.sg \hspace{5pt} xintao.wang@outlook.com\hspace{5pt} gujinwei@sensebrain.ai}\\
}

\thispagestyle{empty}
\twocolumn[{%
			\renewcommand\twocolumn[1][]{#1}%
			\vspace{-1em}
			\maketitle
			\vspace{-1em}
			\begin{center}
				\centering
				\vspace{-0.2in}
				\includegraphics[width=0.99\textwidth]{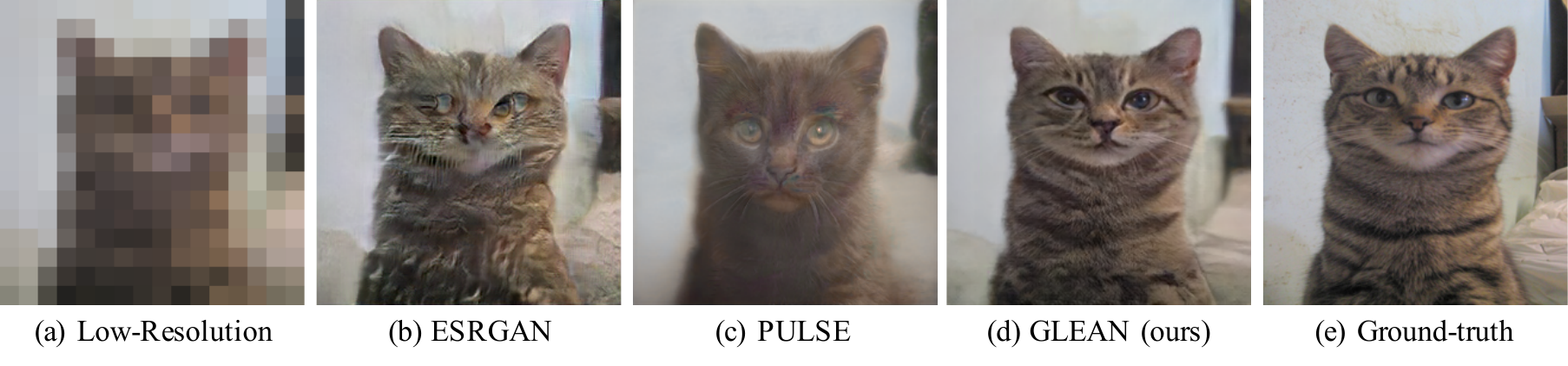}\vspace{0.1cm}
				\vskip -0.4cm
				\captionof{figure}{\textbf{Example of large-factor super-resolution (16${\times}$)}.
					(a) The low-resolution input (LR).
					(b) ESRGAN~\cite{wang2018esrgan} trains the SR generator from scratch, which often produces artifacts and unnatural textures.
					(c) PULSE~\cite{menon2020pulse}  achieves more realistic results by GAN inversion, which, however, cannot faithfully recover the structures of the ground-truth.
					(d) With the proposed generative latent bank, GLEAN is able to generate output that not only is close to the ground-truth, but also possesses realistic textures.
					(e) The ground-truth (GT).
				}
				\label{fig:teaser}
			\end{center}%
		}]

\begin{abstract}
	\vspace{-0.3cm}
	We show that pre-trained Generative Adversarial Networks (GANs), \eg, StyleGAN, can be used as a latent bank to improve the restoration quality of large-factor image super-resolution (SR).\makeatletter{\renewcommand*{\@makefnmark}{}
		\footnotetext{$^*$Corresponding author}\makeatother}
	While most existing SR approaches attempt to generate realistic textures through learning with adversarial loss, our method, \textbf{G}enerative \textbf{L}at\textbf{E}nt b\textbf{AN}k (GLEAN), goes beyond existing practices by directly leveraging rich and diverse priors encapsulated in a pre-trained GAN.
	But unlike prevalent GAN inversion methods that require expensive image-specific optimization at runtime, our approach only needs a single forward pass to generate the upscaled image.
	GLEAN can be easily incorporated in a simple encoder-bank-decoder architecture with multi-resolution skip connections. Switching the bank allows the method to deal with images from diverse categories, \eg, cat, building, human face, and car. Images upscaled by GLEAN show clear improvements in terms of fidelity and texture faithfulness in comparison to existing methods as shown in Fig.~\ref{fig:teaser}. A high-resolution version of this paper can be found at \url{https://ckkelvinchan.github.io/}.
	\vspace{-0.5cm}

\end{abstract}

\section{Introduction}
In this study, we explore a new way to employ GAN~\cite{goodfellow2014generative} for image super-resolution. We are interested in the regime of high magnification factors ($8{\times}$ to ${64\times}$), which typical SR methods fail to handle since most details and textures are lost during downsampling. Since the problem is severely underspecified, informative priors become inevitable in this setting, especially in restoring the textural details.
Studying large-factor image SR is meaningful as it can potentially improve the state of the arts in SR, and more generally conditional generative models for images.

The notion of GAN has been extensively used in SR with the aim to enrich texture details in an upscaled image.
There are two popular approaches to deploy GANs for this task.
The more common paradigm~\cite{ledig2017photo,wang2018recovering,wang2018esrgan} trains a generator to handle the upscaling task, where adversarial training is performed by using a discriminator to differentiate real images from the upscaled images produced by the generator.
Another possible way to exploit GAN for the task is by GAN inversion~\cite{bau2020semantic,gu2020image,menon2020pulse,pang2020exploiting}. In this setting, one will need to `invert' the generation process of a pre-trained GAN by mapping a corrupted image back to the latent space. A restored image can then be reconstructed from the optimal vector in the latent space.

While both methods are capable of generating more realistic results than approaches that solely rely on $\ell_2$ loss, they have some inherent shortcomings.
The first paradigm typically trains the SR generator \textit{from scratch} using a combined objective function consisting of a fidelity term and an adversarial loss.
In this setting, the generator is responsible for both capturing the natural image characteristics and maintaining the fidelity to the ground-truth. This inevitably limits the capability of approximating the natural image manifold. As a result, these methods often produce artifacts and unnatural textures.
As shown in Fig.~\ref{fig:teaser}, while ESRGAN~\cite{wang2018esrgan} faithfully recovers the structures (\eg~pose, ear shape) of the cat, it struggles to produce realistic textures.

The second paradigm resolves the aforementioned problem by making better use of the latent space of GAN through optimization. However, as the low-dimensional latent codes and the constraints in the image space are insufficient to guide the restoration process, these methods often generate images with low fidelity. As shown in Fig.~\ref{fig:teaser}, despite being realistic, the output of a representative method, PULSE~\cite{menon2020pulse}, fails to recover the structures of the ground-truth faithfully.
In addition, as the optimization is usually conducted in an iterative manner for each image at runtime, these approaches are often time-consuming.

In our approach, we leverage pre-trained GANs such as StyleGAN~\cite{karras2018style} to provide rich and diverse priors for the task.
Unlike most GAN inversion methods, which also use pre-trained GANs, our method does not involve image-specific optimization at runtime. Once trained, the model only needs a single forward pass to upscale an image, which is more practical for applications that demand fast response.
The idea is partially inspired by the classic notion of dictionary~\cite{yang2010image}. But unlike conventional approaches that construct a finite and imagery-derived dictionary, we exploit GAN as a more effective way for storing priors.

Conditioning and retrieving from a \textit{GAN-based dictionary} is a new and non-trivial question we need to address in this work. We show that pre-trained GANs can be employed as a latent bank in a succinct \textit{encoder-bank-decoder} architecture.
This novel architecture allows us to lift the burden of learning both fidelity and texture generation simultaneously in a typical encoder-decoder network since the latent bank already captures rich texture priors.
In addition, we show that it is pivotal to condition the bank by passing both the latent vectors and multi-resolution convolutional features from the encoder to achieve high-fidelity results. Symmetrically, multi-resolution cues need to be passed from the bank to the decoder.
We show the effectiveness of the proposed method in handling images with challenging poses and structures apart from the large magnification factor. We also demonstrate how the method can be generalized to different categories, \eg, human faces, cats, buildings, by switching different pre-trained GAN latent banks.


\begin{figure*}[!t]
	\begin{center}
		\includegraphics[width=\textwidth]{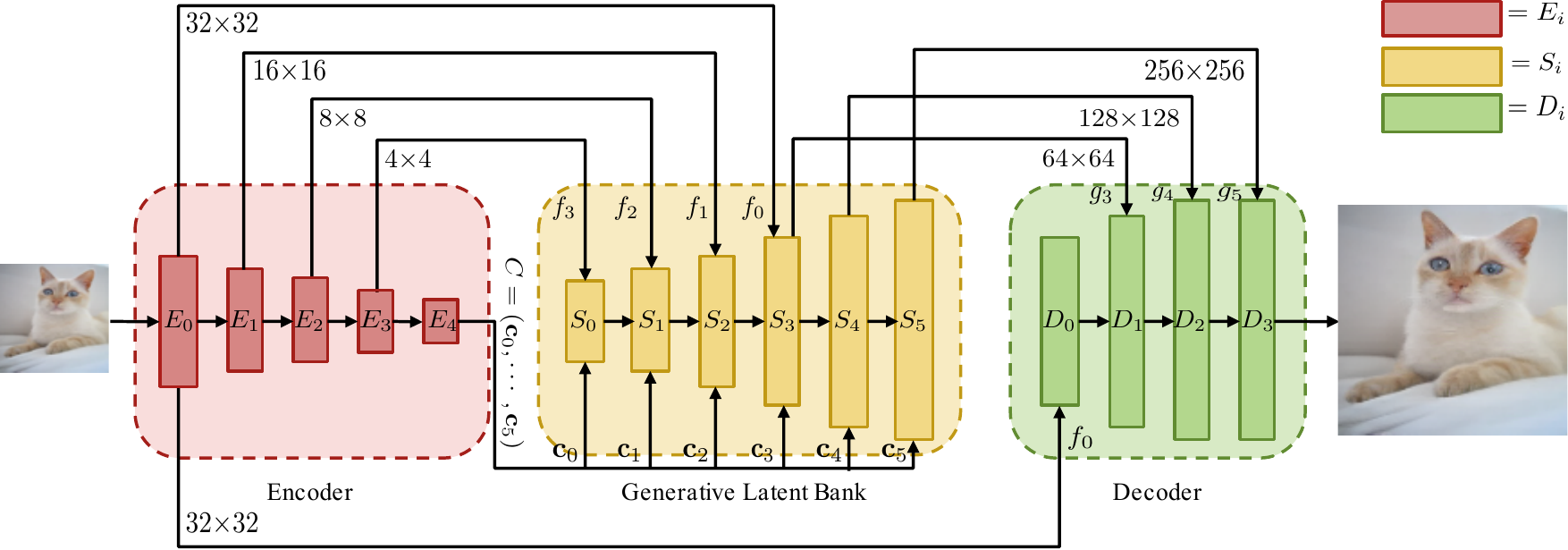}
		\caption{\textbf{Overview of GLEAN}. In addition to the latent vectors $\mathbf{c}_i$, the generator (\ie, the generative latent bank) is also conditioned on the multi-resolution features $f_i$. With a pre-trained GAN capturing the natural image prior, this encoder-bank-decoder design lifts the burden of learning both fidelity and naturalness in the conventional encoder-decoder architecture. $E_i$, $S_i$ and $D_i$ denote the encoder blocks, latent bank blocks and decoder blocks, respectively. This example corresponds to an input size of $32{\times}32$ and an output size of $256{\times}256$.
		}
		\label{fig:overview}
		\vspace{-0.4cm}
	\end{center}
\end{figure*}
\section{Related Work}
\label{sec:relatedwork}

\noindent{\textbf{Image Super-Resolution.}}
Many existing SR algorithms~\cite{dai2019secondorder,dong2014learning,dong2016image,dong2016accelerating,he2019ode,xu2019towards,zhang2018image,zhou2020cross} directly learn a mapping from the low-resolution images to high-resolution images with a pixel-wise constraint (\eg $\ell_2$ loss).
While these methods achieve remarkable results in terms of PSNR, training solely with pixel-wise constraints often results in perceptually unconvincing outputs with severe over-smoothing artifacts~\cite{ledig2017photo,menon2020pulse}.
To alleviate the problem, GANs~\cite{ledig2017photo,sajjadi2017enhancenet,wang2018esrgan,xu2017learning} are employed to approximate the natural image manifold, yielding more photo-realistic results. However, as the generator needs to learn both fidelity and natural image characteristics, unnatural artifacts could still be observed in the outputs, especially if one trains the generator from scratch.

Recent interests have shifted to large-factor SR beyond the typical upscaling factors ($2{\times}$ or $4{\times}$)~\cite{dahl2017pixel,hyun2020varsr,shang2020perceptual,zhang2019texture}.
Dahl~\etal~\cite{dahl2017pixel} propose a fully probabilistic pixel recursive network for upsampling extremely coarse images with resolution $8{\times}8$.
RFB-ESRGAN~\cite{shang2020perceptual} builds upon ESRGAN and adopts multi-scale receptive fields blocks for $16{\times}$ SR.
VarSR~\cite{hyun2020varsr} achieves $8{\times}$ SR by matching the latent distributions of LR and HR images to recover the missing details.
Zhang~\etal~\cite{zhang2019texture} perform $16{\times}$ reference-based SR on paintings with a non-local matching module and a wavelet texture loss.
To handle even larger magnification factors, one would need to rely on stronger priors. SR methods specialized on large magnification factors are typically dedicated to the human face category as one could exploit the strong structural prior of faces.
Facial priors including facial attributes~\cite{li2019deep}, facial landmarks~\cite{kim2019progressive,ma2020deep}, and identity~\cite{grm2019face} have been studied.
Our work goes beyond previous works and pushes the limit to $64{\times}$ and generalizes to more categories. Such a large magnification factor is challenging due to its highly ill-posed nature.

\noindent{\textbf{GAN Inversion.}}
Given a degraded image $x$, GAN inversion-based methods~\cite{bau2020semantic,gu2020image,menon2020pulse,pang2020exploiting} in general produce a natural image best approximating $x$ by optimizing $z^* = \argmin\nolimits_{z \in \mathcal{Z}} \mathcal{L}\left(G(z), x\right)$, where $\mathcal{Z}$ is the latent space and $\mathcal{L}(\cdot,\cdot)$ denotes the task-specific objective function.
For instance, PULSE~\cite{menon2020pulse} iteratively optimizes the latent code of StyleGAN~\cite{karras2018style} with a pixel-wise constraint between the input and output.
mGANprior~\cite{gu2020image} optimizes multiple latent codes to increase the expressiveness of the model.
DGP~\cite{pang2020exploiting} further finetunes the generator together with the latent code to reduce the gap between the distributions of the training and testing images.
A common issue with GAN inversion is that important spatial information may not be faithfully kept due the low-dimensionality of the latent code. Thus, these methods often generate undesirable results that do not resemble the ground-truth.
Different from GAN inversion, GLEAN conditions the pre-trained generator with both the latent codes and multi-resolution convolutional features, providing additional spatial guidance for restoration. In addition, GLEAN does not require iterative optimization during inference.


\section{Methodology}
\label{sec:method}
A GAN model that is trained on large-scale natural images captures rich texture and shape priors.
Previous studies~\cite{bau2020semantic,gu2020image,menon2020pulse,pang2020exploiting} have shown that such priors can be harvested through GAN inversion to benefit various image restoration tasks. Nonetheless, it remains underexplored how to exploit the priors without the expensive optimization during inversion.

In this study, we devise GLEAN within a novel \textit{encoder-bank-decoder} architecture, which allows one to exploit the generative priors by needing just a single forward pass. An overview of the architecture is depicted in Fig.~\ref{fig:overview}.
Given a severely downsampled LR image, GLEAN applies an encoder to extract latent vectors and multi-resolution convolutional features, which capture important high-level cues as well as spatial structure of the LR image.
Such cues are used to condition the latent bank, which further produces another set of multi-resolution features for the decoder.
Finally, the decoder generates the final output by integrating the features from both the encoder and the latent bank.
In this work, we adopt StyleGAN~\cite{karras2018style,karras2019analyzing} as the generative latent bank due to its exceptional performance. The idea of latent bank can be extended to other generators such as BigGAN~\cite{brock2018large}.

\subsection{Encoder}

To generate the latent vectors, we first use an RRDBNet~\cite{wang2018esrgan} (denoted as $E_0$) to extract features $f_0$ from the input LR image.
Then, we gradually reduce the resolution of the features by:
\begin{equation}
	f_i = E_i(f_{i-1}),\quad i\in\{1, \cdots, N\},
\end{equation}
where $E_i$, ${i\in\{1, \cdots, N\}}$, denotes a stack of a stride-2 convolution and a stride-1 convolution. Finally, a convolution and a fully-connected layer are used to generate the latent vectors:
\begin{equation}
	C = E_{N + 1}(f_{N}),
\end{equation}
where $C$ is a matrix whose columns represent the latent vectors for the StyleGAN.

The latent vectors in $C$ capture a compressed representation of the images, providing the generative latent bank with high-level information.
To further capture the local structures of the LR image and to provide additional guidance for structure restoration, we also feed multi-resolution convolutional features $\{f_i\}$ into the latent bank.

\subsection{Generative Latent Bank}
Given the convolutional features $\{f_i\}$ and the latent vectors $C$, we leverage a pre-trained generator as a latent bank to provide priors for texture and detail generation.
As StyleGAN is originally designed for image generation tasks, it cannot be directly integrated into the proposed encoder-bank-decoder framework.
In this work, we adapt StyleGAN to our SR network by making three modifications:
\begin{enumerate}
	\item Instead of taking one single latent vector as the input, each block of the generator takes a different latent vector to improve expressiveness. More specifically, we have $C{=} (\mathbf{c}_0, \cdots, \mathbf{c}_{k-1})$ for $k$ blocks, where each $\mathbf{c}_i$ corresponds to one latent vector. We find that this modification leads to outputs with fewer artifacts. This modification is also seen in previous works~\cite{gu2020image,xu2020generative,zhu2020in}.
	\item To allow conditioning on the additional features from the encoder, we use an additional convolution in each style block for feature fusion:
	      \begin{equation}
		      g_i =
		      \begin{cases}
			      S_0(\mathbf{c}_0, f_{N}),              & \text{if }i = 0,  \\
			      S_i(\mathbf{c}_{i}, g_{i-1}, f_{N-i}), & \text{otherwise},
		      \end{cases}
	      \end{equation}
	      where $S_i$ denotes the augmented style block with an additional convolution, and $g_i$ corresponds to the output feature of the $i$-th augmented style block.
	\item Instead of directly generating outputs from the generator, we output the features $\{g_i\}$ and pass them to the decoder to better fuse the features from the latent bank and encoder.
\end{enumerate}
\noindent{\textbf{Advantages.}} The use of generative latent bank is reminiscent of the task of reference-based SR~\cite{li2020blind,li2020enhanced,yan2020towards,zhang2020copy,zhang2019image}, where external HR reference image(s) are employed as an explicit imagery dictionary.
While the external HR information leads to marked improvements, the performance is sensitive to the similarity between the inputs and references. This sensitivity may eventually lead to degraded results when the reference images/components are not well selected.
Moreover, the size and diversity of those imagery dictionaries are limited by the selected components, impeding the generalization to diverse scenes in practice.
In addition, computationally-intensive global matching~\cite{zhang2019image} or component detection/selection~\cite{li2020blind} is often required to aggregate appropriate information from the references, hindering the applications to scenarios with tight computational constraints.
Instead of constructing an imagery dictionary, GLEAN adopts a \textit{GAN-based} dictionary conditioned on a pre-trained GAN. Our dictionary does not depend on any specific components or images. Instead, it captures the distribution of the images and has potentially unlimited size and diversity.
Furthermore, GLEAN is computationally efficient without requiring global matching and reference images/components selection.

\subsection{Decoder}
GLEAN uses an additional decoder with progressive fusion to integrate the features from the encoder and latent bank to generate the output image.
It takes the RRDBNet features as inputs and progressively fuse the features with the multi-resolution features from the latent bank:
\begin{equation}
	d_i =
	\begin{cases}
		D_0(f_0)                    & \text{if } i = 0, \\
		D_i(d_{i-1}, g_{N - 2 + i}) & \text{otherwise},
	\end{cases}
\end{equation}
where $D_i$ and $d_i$ denote a $3{\times}3$ convolution and its output, respectively. Each convolution is followed by a pixel-shuffle~\cite{shi2016real} layer except the final output layer.
With the skip-connection between the encoder and decoder, the information captured by the encoder can be reinforced and hence the latent bank could focus more on the texture and detail generation.

\subsection{Training}
Similar to existing works~\cite{ledig2017photo,wang2018recovering,wang2018esrgan}, we adopt the standard $l_2$ loss, perceptual loss~\cite{johnson2016perceptual}, and adversarial loss for training. More details on the loss function can be found in the appendix.
To exploit the generative prior, we keep the weights of the latent bank fixed throughout training.
In our preliminary experiments, finetuning the latent bank with the encoder and decoder demonstrates no noticeable improvements. Moreover, it potentially harms the generalizability of the model as the latent bank may eventually bias to the training distribution.
It is worth emphasizing that despite GLEAN is trained with similar objectives as in existing works (\eg~ESRGAN), the main difference to these methods is that GLEAN leverages a pre-trained generator to directly incorporate the priors into the network, further improving the output quality. We show that the improvement is not due to additional parameters in the generator by comparing GLEAN with ESRGAN$^+$, a larger ESRGAN that has similar FLOPs to GLEAN.

\begin{figure*}[!t]
    \begin{center}
        \includegraphics[width=0.99\textwidth]{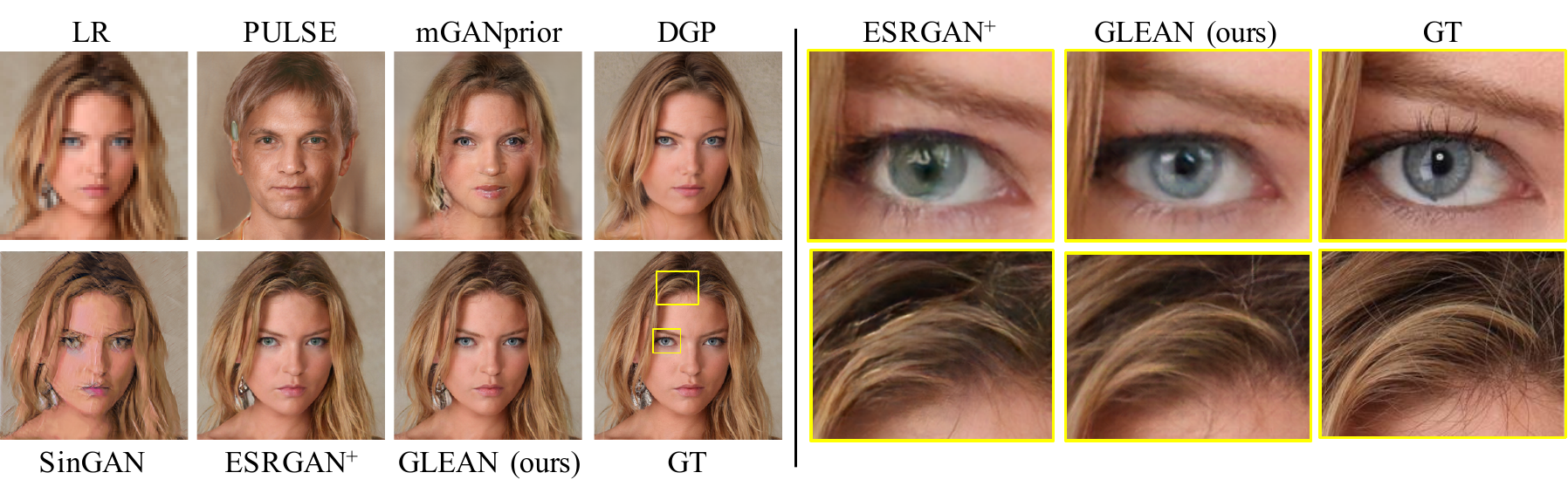}
        \caption[]{\textbf{Comparisons on 16${\times}$ SR on CelebA-HQ~\cite{karras2018progressive}.} Only GLEAN is able to maintain high fidelity while synthesizing realistic textures and details: GAN inversion methods fail to preserve the identity, and adversarial loss methods struggle to synthesize fine details. ESRGAN$^{+}$ denotes a larger version with similar FLOPs to GLEAN. \textbf{(Zoom-in for best view)}}
        \label{fig:sr}
        \vspace{-0.3cm}
    \end{center}
\end{figure*}
\begin{figure}[!t]
    \begin{center}
        \includegraphics[width=0.49\textwidth]{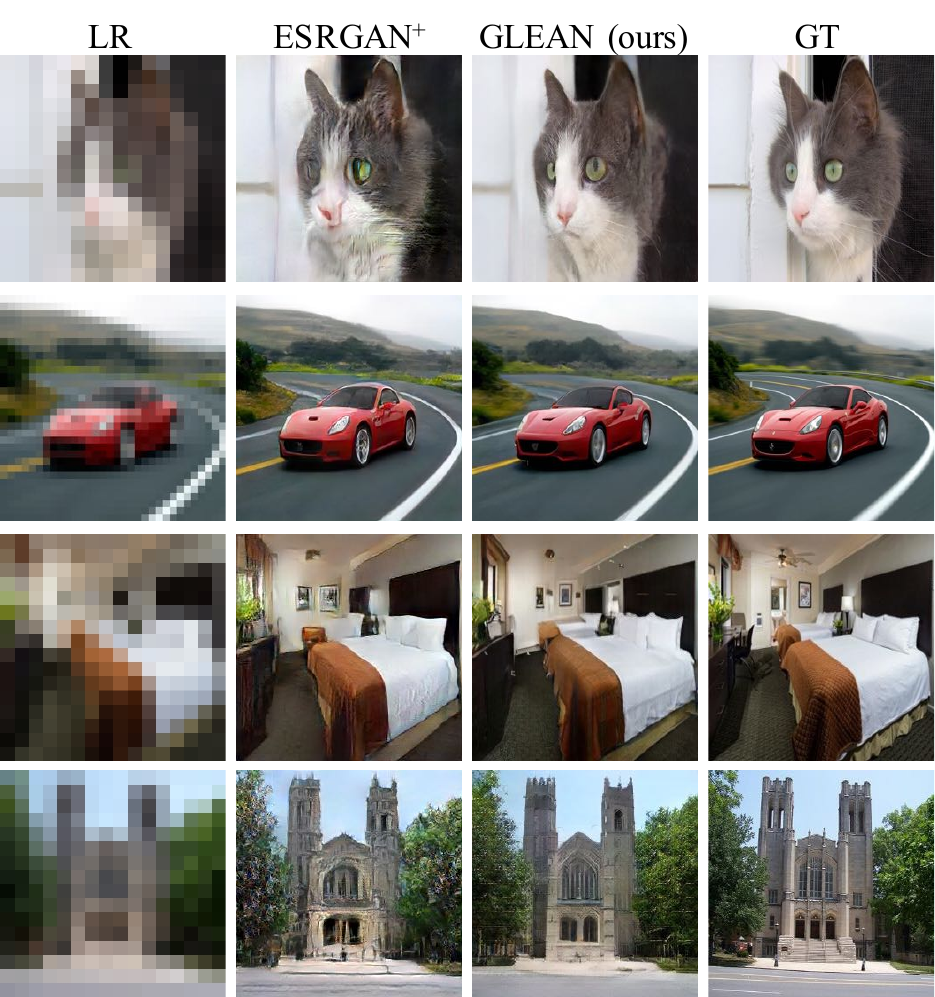}
        \caption{\textbf{Results of 16${\times}$ SR on other categories.} GLEAN can be applied to various categories by switching between StyleGANs trained on different categories. \textbf{(Zoom-in for best view)}}
        \label{fig:sr_others}
    \end{center}
    \vspace{-0.45cm}
\end{figure}
\begin{figure}[!t]
    \begin{center}
        \includegraphics[width=0.49\textwidth]{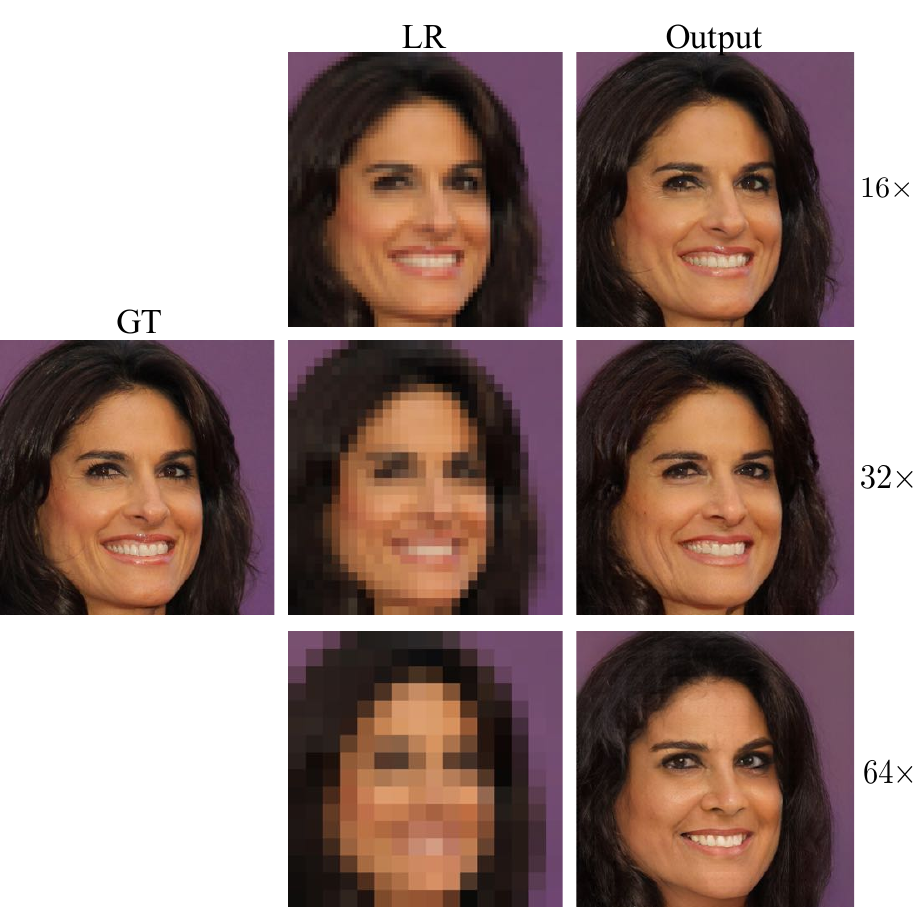}
        \caption{\textbf{Results on larger scale factors.} GLEAN reconstructs realistic images highly similar to the GT for up to $64{\times}$ upscaling factor. \textbf{(Zoom-in for best view)}}
        \label{fig:sr_scale}
    \end{center}
\end{figure}

\section{Experiments}
\label{sec:exp}
We adopt pre-trained StyleGAN\footnote{GenForce: https://github.com/genforce/genforce}~\cite{karras2018style} or StyleGAN2\footnote{BasicSR: https://github.com/xinntao/BasicSR}~\cite{karras2019analyzing} (depending on the availability of pre-trained models) as our latent bank, and use the publicly available codes of existing methods for the comparison in this section.
To maintain fairness, we train our model and baselines on the same datasets, including FFHQ~\cite{karras2018style} and LSUN~\cite{yu2015lsun}, so that the difference in restoration quality is mainly caused by the algorithms instead of the training distribution. Test set is strictly exclusive from the training. Detailed experimental settings are provided in the appendix.
\vspace{0.15cm}

\noindent{\textbf{Qualitative comparison.}}
The qualitative comparison on $16{\times}$ SR is shown in Fig.~\ref{fig:sr}. Guided by low-dimensional vectors and constraints in LR space, the outputs of GAN inversion methods are unable to maintain a good fidelity. In particular, PULSE~\cite{menon2020pulse} and mGANprior~\cite{gu2020image} fail to restore a face image with the same identity. In addition, artifacts are observed in their outputs.
Through finetuning the generator during optimization, the result of DGP~\cite{pang2020exploiting} demonstrates significant improvements in both quality and fidelity. However, a slight difference between the identities of the output and ground-truth is still observed. For example, the eyes and lips show noticeable differences.

Methods trained with adversarial loss (SinGAN~\cite{shaham2019singan}, ESRGAN$^+$\footnote{A larger version of ESRGAN with similar FLOPs to GLEAN.}~\cite{wang2018esrgan}) can preserve the local structures, but fail in synthesizing convincing textures and details.
Specifically, SinGAN fails to capture the natural image style, producing a painting-like image. Although ESRGAN$^+$ is capable of generating a realistic image, it struggles to synthesize fine details and introduces unnatural artifacts in detailed regions. It is worth emphasizing that although ESRGAN$^+$ achieves competitive results on human faces, its performances on other categories such as \textit{cats} and \textit{cars} are less promising (see Fig.~\ref{fig:teaser} and Fig.~\ref{fig:sr_others}).
With the latent bank providing natural image priors, GLEAN succeeds in both fidelity and naturalness.
For example, when compared to ESRGAN$^+$, GLEAN reconstructs eyes with better shape and details.
We further extend our method to larger scale factors in Fig.~\ref{fig:sr_scale}. GLEAN successfully generates perceptually convincing images resembling the ground-truth for up to $64{\times}$ upscaling.
\begin{figure}[!t]
    \begin{center}
        \includegraphics[width=0.49\textwidth]{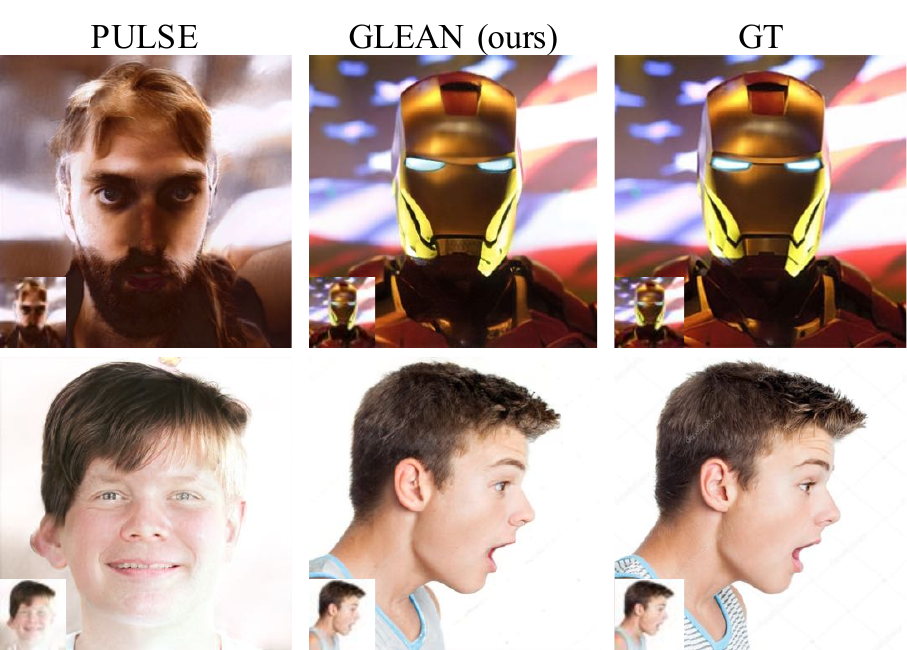}
        \caption{\textbf{Outputs with diverse poses and contents.} Despite GLEAN is trained with aligned human faces, it is able to reconstruct faithful images for non-aligned and non-human faces. PULSE approximates the GT in low resolution \textit{(bottom left)}, but its outputs are significantly different from the GT when viewed in high resolution.}
        \label{fig:sr_pose}
    \end{center}
\end{figure}
\begin{table}[!t]
    \caption{\textbf{Cosine similarity of ArcFace Embeddings~\cite{deng2019arcface}.} GLEAN achieves a higher similarity than baselines. \textbf{Bolded} texts represent the best performance.}
    \label{tab:arcface}
    \begin{center}
        \tabcolsep=0.15cm
        \vspace{-0.5cm}
        \scalebox{0.95}{
            \begin{tabular}{l|c|c|c}
                           & PULSE~\cite{menon2020pulse}    & mGANprior~\cite{gu2020image}     & DGP~\cite{pang2020exploiting} \\\hline
                Similarity & 0.4047                         & 0.5526                           & 0.7341                        \\ \hline\hline
                           & SinGAN~\cite{shaham2019singan} & ESRGAN$^+$~\cite{wang2018esrgan} & \textbf{GLEAN}                \\\hline
                Similarity & 0.7718                         & 0.9599                           & \textbf{0.9678}
            \end{tabular}}
    \end{center}
\end{table}
\begin{table}[!t]
    \caption{\textbf{Quantitative (PSNR/LPIPS) comparison on 16${\times}$ SR.} GLEAN outperforms other methods in most categories. ESRGAN$^+$ denotes a larger version of ESRGAN~\cite{wang2018esrgan} having similar FLOPs to GLEAN. \textbf{Bolded} texts represent the best performance.}
    \label{tab:quan}
    \begin{center}
        \tabcolsep=0.1cm
        \vspace{-0.5cm}
        \scalebox{0.75}{
            \begin{tabular}{l|c|c|c|c}
                                                  & mGANprior~\cite{gu2020image} & PULSE~\cite{menon2020pulse} & ESRGAN$^+$~\cite{wang2018esrgan} & \textbf{GLEAN}                 \\\hline
                Face~\cite{karras2018progressive} & 23.66/0.4661                 & 21.83/0.4600                & 26.76/0.2787                     & \textbf{26.84}/\textbf{0.2681} \\
                Cat~\cite{zhang2008cat}           & 17.01/0.5556                 & 19.78/0.5241                & 19.99/0.3482                     & \textbf{20.92}/\textbf{0.3215} \\
                Car~\cite{krause2013object}       & 14.53/0.7228                 & 16.30/0.6491                & 19.42/0.3006                     & \textbf{19.74}/\textbf{0.2830} \\
                Bedroom~\cite{yu2015lsun}         & 16.38/0.5439                 & 12.97/0.7131                & \textbf{19.47}/\textbf{0.3291}   & 19.44/0.3310                   \\
                Tower~\cite{yu2015lsun}           & 15.96/0.4870                 & 13.62/0.7066                & 17.86/0.3132                     & \textbf{18.41}/\textbf{0.2850} \\
            \end{tabular}}
        \vspace{-0.5cm}
    \end{center}
\end{table}
\begin{figure}[!t]
    \begin{center}
        \hspace{-0.5cm}\includegraphics[width=0.49\textwidth]{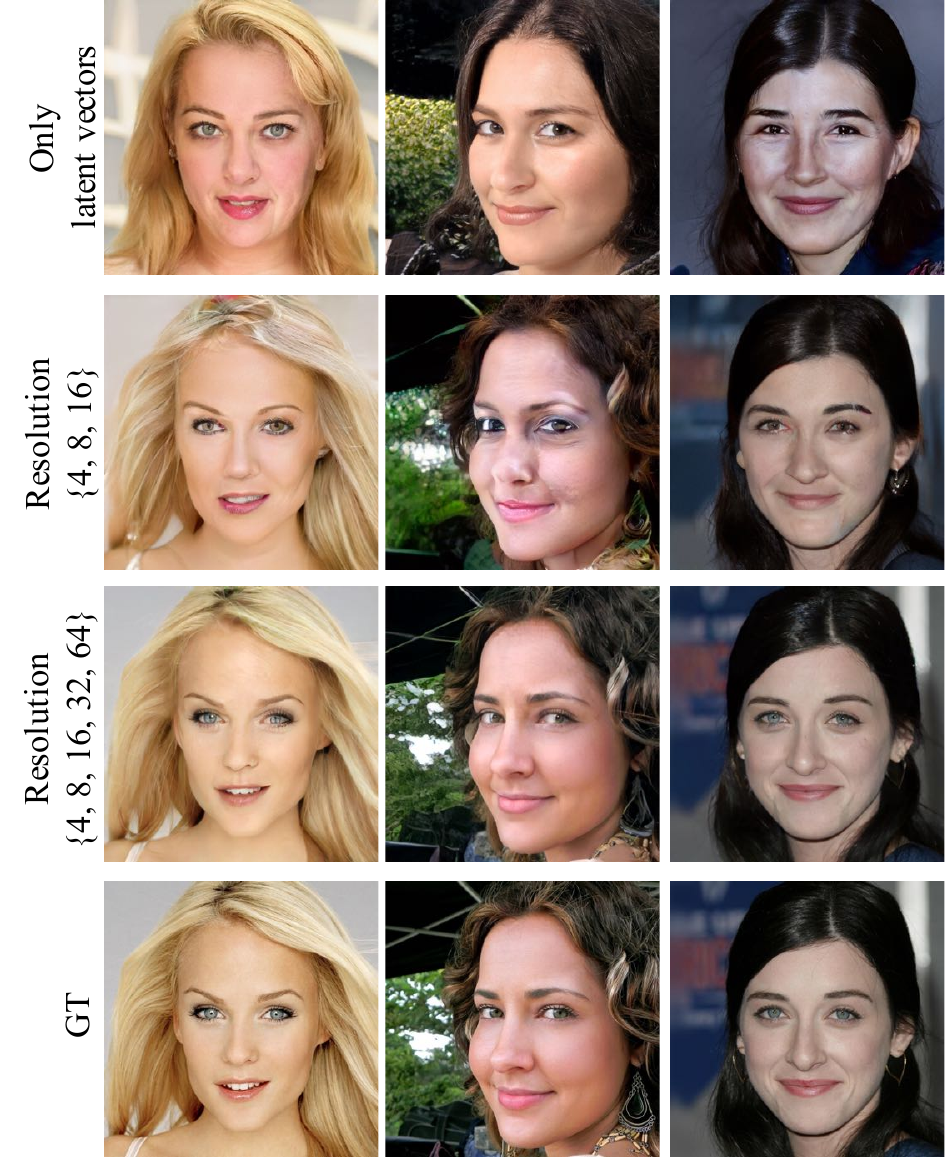}
        \caption{\textbf{Effects of the multi-resolution encoder features.} Without the convolutional features, the outputs can only resemble the global attributes (\eg~hair color, pose). When adding the encoder features progressively, the network can capture more local structures, better approximating the GT. }
        \label{fig:ablation_enc}
        \vspace{-0.6cm}
    \end{center}
\end{figure}
\vspace{0.15cm}

\noindent{\textbf{Robustness to poses and contents.}}
Another appealing property of GLEAN is its robustness to the changes in poses and contents. As shown in Fig.~\ref{fig:sr_pose}, guided by the convolutional features, GLEAN is still able to construct realistic images when the images are non-aligned and contain non-human faces despite it is trained on aligned human faces. In contrast, the outputs of PULSE are biased to aligned human faces. Its outputs can only approximate the ground-truths in low resolution.
Such robustness enables GLEAN to be applied to diverse categories and scenes such as cats, cars, bedrooms, and towers. Examples are shown in Fig.~\ref{fig:sr_others} and more results are provided in the appendix.\vspace{0.3cm}

\noindent{\textbf{Quantitative comparison.}}
To demonstrate the ability of GLEAN in producing outputs with high fidelity,
we extract 100 images from CelebA-HQ~\cite{karras2018progressive} and compute the cosine similarity to the ground-truth on the ArcFace embedding space~\cite{deng2019arcface}. As shown in Table~\ref{tab:arcface},  GLEAN achieves higher similarity than the baseline methods, validating the
superiority of GLEAN.

We additionally provide the quantitative comparison on different categories in Table~\ref{tab:quan}. For each category, we select 100 images and compute their average PSNR and LPIPS~\cite{zhang2018unreasonable}. It is observed that mGANprior and PULSE perform significantly worse as they fail to restore the original objects. GLEAN outperforms these methods in most categories, suggesting its effectiveness in generating images with high quality and fidelity.

\begin{figure}[!t]
    \begin{center}
        \includegraphics[width=0.49\textwidth]{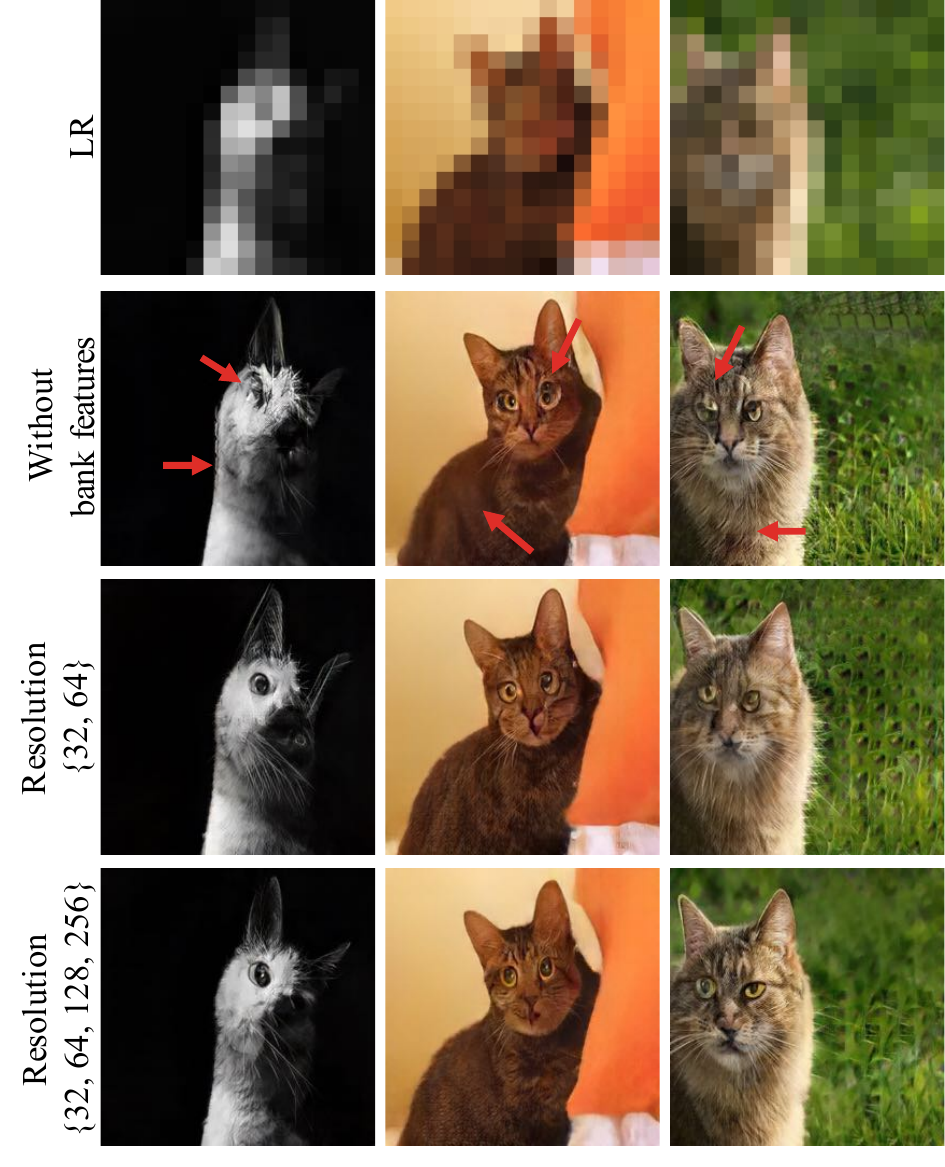}
        \caption{\textbf{Effects of the latent bank features.} The rich texture priors captured in the generator lift the burden of the encoder in texture generation. Improvements on both texture and structures are observed when finer features are inserted into the decoder. \textbf{(Zoom-in for best view)}}
        \vspace{-0.5cm}
        \label{fig:ablation_gen}
    \end{center}
\end{figure}
\begin{figure}[!t]
    \begin{center}
        \includegraphics[width=0.49\textwidth]{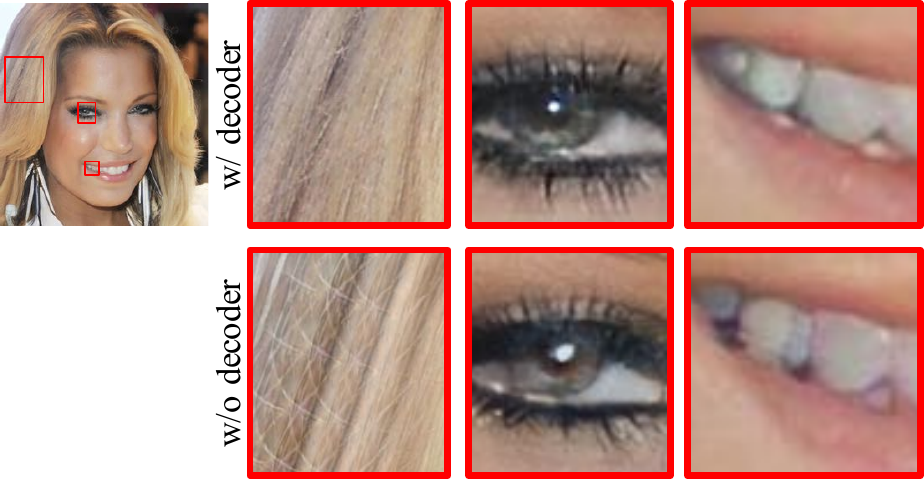}
        \caption{\textbf{Contributions of the decoder.} The decoder reinforces the spatial information captured in the encoder features and aggregate them in a coarse-to-fine manner, leading to enhanced quality.}
        \vspace{-0.5cm}
        \label{fig:ablation_dec}
    \end{center}
\end{figure}

\section{Ablation Studies}
\label{sec:ablation_enc}
\noindent\textbf{Importance of multi-resolution encoder features.}
We demonstrate how the convolutional features generated from the encoder assist in the restoration of fine details and local structures.
We start with only the latent vectors and observe the transition when features are gradually introduced to the latent bank as conditions.
To discard the effects brought by the decoder, we test with a variant of GLEAN where the generator directly produces the output images. The comparison is depicted in Fig.~\ref{fig:ablation_enc}.

When all convolutional features are discarded, GLEAN resembles the typical GAN inversion methods that learn only the latent vectors. Similar to those methods, the network is able to synthesize realistic images given the latent vectors. However, guided only by low-dimensional vectors, in which spatial information is not well-preserved, the network restores only the global attributes such as hair color and poses, but fails to preserve finer details.
When providing coarse (from $4{\times}4$ to $16{\times}16$) convolutional features to the latent bank, more details are recovered and the outputs are better approximating the ground-truths. Further improvements in both quality and fidelity are observed when finer features are passed to the latent bank.
The above observations corroborate our hypothesis that the convolutional features are pivotal in guiding the restoration of fine details and local structures, which cannot be reconstructed with only the latent vectors.
\vspace{0.15cm}

\noindent\textbf{Effects of latent bank features.}
To understand the contributions of the latent bank, we investigate the effects brought by the latent bank features. We start by discarding all the latent bank features, and progressively pass the features to the decoder. The comparison is shown in Fig.~\ref{fig:ablation_gen}.
Lacking appropriate prior information, the network is responsible for both generating realistic details and maintaining fidelity to the ground-truths. Such a demanding objective eventually leads to outputs that contain flaws in both structure restoration and texture generation.
With the latent bank, the burden of texture and details generation is reduced as the generator already captures rich image priors. Therefore, improvements in both structures and textures are observed when passing finer features to the decoder.
\vspace{0.15cm}

\noindent\textbf{Importance of decoder.}
As shown in Fig.~\ref{fig:ablation_dec}, without the decoder, despite being perceptually convincing overall, the output image contains unpleasant artifacts when zoomed in.
The decoder allows the network to aggregate the information in a coarse-to-fine manner, leading to more natural details. In addition, the multi-scale skip-connections between the encoder and decoder reinforce the spatial information captured in the encoder features so that the latent bank could focus more on detail generation, further enhancing the output quality.
\begin{figure}[!t]
    \begin{center}
        \subfigure[Comparison with DFDNet~\cite{li2020blind}]{\includegraphics[width=0.48\textwidth]{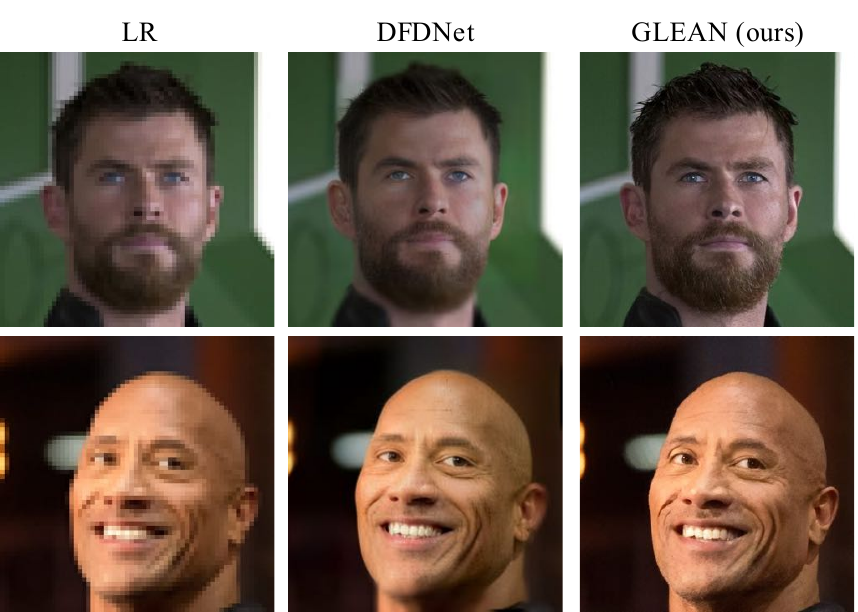}}
        \vspace{-0.1cm}
        \subfigure[Comparison with SRNTT~\cite{zhang2019image}]{\includegraphics[width=0.48\textwidth]{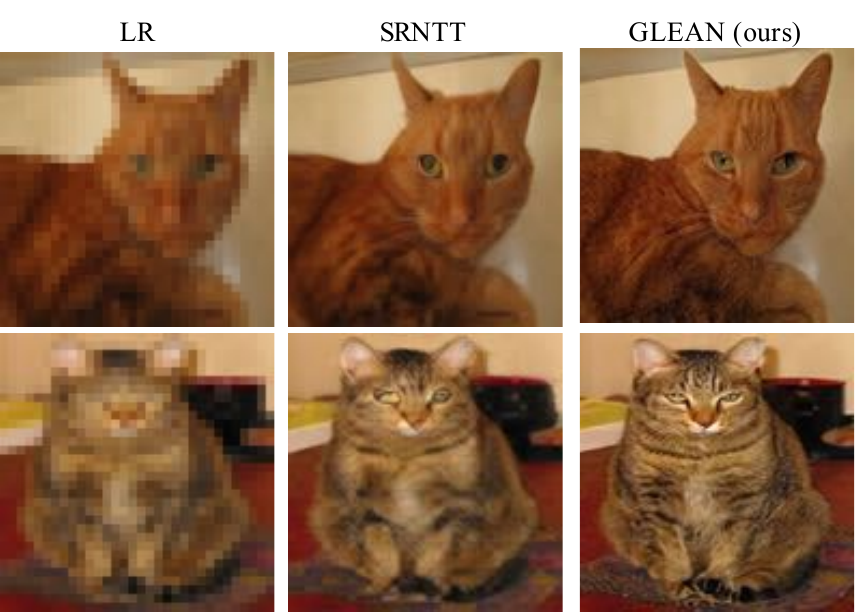}}
        \caption{\textbf{Comparison to imagery dictionary.} \textbf{(a)} DFDNet fails to restore components absent in the dictionary (\eg~skin, hair), leading to incoherent outputs. \textbf{(b)} SRNTT is unable to produce faithful fur textures.}
        \label{fig:ablation_dict}
        \vspace{-0.7cm}
    \end{center}
\end{figure}
\vspace{0.15cm}

\noindent\textbf{Comparisons with reference-based methods.}
We assess the efficacy of the new notion of GAN-based dictionary by comparing GLEAN with two representative methods adopting an imagery dictionary for SR -- SRNTT~\cite{zhang2019image} and DFDNet~\cite{li2020blind}. Examples are shown in Fig.~\ref{fig:ablation_dict}.

For DFDNet, we evaluate the performance on LR images with unknown degradations\footnote{We further downsample the LR images to $64{\times}64$ to match the input size of GLEAN.}. Through pre-constructing a dictionary of facial components (\eg~eyes, lips), DFDNet shows remarkable performance on face restoration. However, it cannot produce faithful results on parts absent in the dictionary, such as skin and hair. Therefore, significant incoherence is observed in the outputs.
Despite GLEAN is trained on the bicubic kernel, it is still capable of producing appealing outputs. More importantly, GLEAN is not confined to improving the visual quality of specific components. Instead, the entire image is super-resolved, leading to coherent and pleasing results. The performance of GLEAN could be further improved by employing multiple degradations during training.

For SRNTT, we follow the same settings and downsample the ground-truth images using the bicubic kernel. With such low-resolution images ($32{\times}32$), global matching becomes prohibitive, and hence SRNTT fails to transfer the textures from HR reference images. As a result, SRNTT tends to provide blurry textures. By capturing the distribution instead of specific imagery clues, GLEAN does not rely on any explicit textural transferal procedure. This enables the applicability to large-factor SR, where image matching is extremely difficult.
More importantly, with no external images employed, GLEAN does not require any global matching to search for suitable textures/details. This allows GLEAN to be applied to images with larger resolutions, where global matching is computationally prohibitive.
\begin{figure}[!t]
    \begin{center}
        \includegraphics[width=0.49\textwidth]{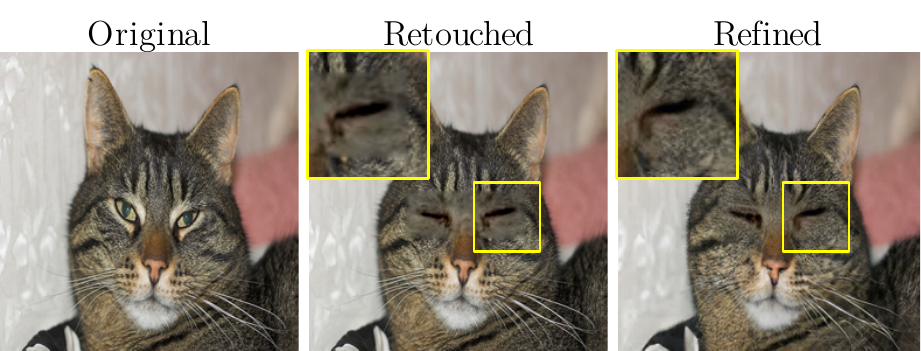}
        \caption{\textbf{Image Retouching.} GLEAN can be used to eliminate unnatural artifacts introduced by amateur retouching. \textbf{(Zoom-in for best view)}}
        \vspace{-0.5cm}
        \label{fig:retouch}
    \end{center}
\end{figure}
\section{Application -- Image Retouching}
In this section, we present one interesting application of GLEAN.
In interactive image retouching, users can manually edit the images based on their preference. However, a perfect output requires tedious and precise retouching. As a result, artifacts are common in the outputs, especially those from amateur retouching.
As a powerful super-resolver, GLEAN can be used as an image retouching tool to eliminate unpleasant artifacts.

As shown in Fig.~\ref{fig:retouch}, the blending operation in the interactive editing software produces a blurry and incoherent output.
Thanks to the capability of GLEAN in producing high quality and fidelity images, GLEAN is able to eliminate the blurry region and generate a coherent output with natural textures.
Furthermore, with only a single forward pass for generation, it can be easily incorporated into existing interactive editing software. More examples will be shown in the appendix.

\section{Conclusion}
We have presented a new way to exploit pre-trained GANs for the task of large-scale super-resolution, up to $64{\times}$ upscaling factor. We have shown that a pre-trained GAN can be used as a generative latent bank in an encoder-bank-decoder architecture. Reconstructing photorealistic HR images requires just a single forward pass, thanks to effective ways in conditioning and retrieving rich priors from the bank. The generality of the notion of GAN-based dictionary allows GLEAN to be potentially extended to not only diverse architectures but also various imaging tasks, such as image denoising, inpainting and colorization.
\clearpage

\noindent\textbf{Acknowledgement}.
This research was conducted in collaboration with SenseTime and supported by the Singapore Government through the Industry Alignment Fund - Industry Collaboration Projects Grant. It is also partially supported by Singapore MOE AcRF Tier 1 (2018-T1-002-056) and NTU SUG.

{\small
	\bibliographystyle{ieee_fullname}
	\bibliography{short,bib.bib}
}
\clearpage
\appendix
\noindent\textbf{\large{Appendix}}\vspace{0.15cm}
\label{appendix}

\textit{We first provide the implementation details of GLEAN in Sec.~\ref{apx:training}}. We then provide additional qualitative results on various categories and scale factors in Sec.~\ref{apx:quali}. Finally, we demonstrate the application of GELAN to the task of image retouching in Sec.~\ref{apx:retouch}.

\section{Training Details of GLEAN}
\label{apx:training}
We adopt pre-trained StyleGAN\footnote{GenForce: https://github.com/genforce/genforce}~\cite{karras2018style} or StyleGAN2\footnote{BasicSR: https://github.com/xinntao/BasicSR}~\cite{karras2019analyzing} as our generative latent bank.
In this section, we assume the latent bank is pre-trained and present the training details of GLEAN (\ie~the encoder-bank-decoder network). Note that the weights of the latent bank are fixed when training GLEAN to better employ the generative prior and to avoid biasing to the training distribution.

We train GLEAN on five categories including human faces, cats, cars, towers, and bedrooms. The training and test datasets used in our experiments are summarized in Table~\ref{tab:dataset}. Since StyleGAN produces images with fixed size, we resize the images in the datasets for our experiments.
\begin{table}[!h]
	\caption{\textbf{Datasets used in our experiments.}}
	\label{tab:dataset}
	\begin{center}
		\tabcolsep=0.15cm
		\vspace{-0.5cm}
		\scalebox{1}{
			\begin{tabular}{l|c|c}
				                     & \textbf{Train}               & \textbf{Test}                          \\\hline
				\textbf{Human faces} & FFHQ~\cite{karras2018style}  & CelebA-HQ~\cite{karras2018progressive} \\
				\textbf{Cats}        & LSUN-train~\cite{yu2015lsun} & CAT~\cite{zhang2008cat}                \\
				\textbf{Cars}        & LSUN-train~\cite{yu2015lsun} & Cars~\cite{krause2013object}           \\
				\textbf{Bedrooms}    & LSUN-train~\cite{yu2015lsun} & LSUN-validate~\cite{yu2015lsun}        \\
				\textbf{Towers}      & LSUN-train~\cite{yu2015lsun} & LSUN-validate~\cite{yu2015lsun}        \\
			\end{tabular}}
	\end{center}
\end{table}

Following previous works~\cite{wang2018recovering,wang2018esrgan}, the objective function for GLEAN consists of three terms. MSE loss is used to guide the fidelity of the output images:
\begin{equation}
	\mathcal{L}_{mse} = \dfrac{1}{N}||\hat{y} - y||_2^2,
\end{equation}
where $N$, $\hat{y}$, and $y$ denote the number of pixels, the output image, and the ground-truth image, respectively. We further incorporate perceptual loss~\cite{johnson2016perceptual} and adversarial loss~\cite{goodfellow2014generative} to improve the perceptual quality:
\begin{align}
	 & \mathcal{L}_{percep} = \dfrac{1}{N}||f(\hat{y}) - f(y)||_2^2, \\
	 & \mathcal{L}_{gen} = \log \left(1 - D(\hat{y})\right),
\end{align}
where $f(\cdot)$ denotes the feature embedding space of the VGG16~\cite{simonyan2014very} network, and $D$ corresponds to the StyleGAN discriminator. The resulting objective function is a weighted mean of the three losses:
\begin{equation}
	\mathcal{L}_g = \mathcal{L}_{mse} + \alpha_{percep}{\cdot}\mathcal{L}_{percep} + \alpha_{gen}{\cdot}\mathcal{L}_{gen}.
\end{equation}
In all our experiments, we set $\alpha_{percep}{=}\alpha_{gen}{=}10^{-2}$. For the discriminator, we maximize
\begin{equation}
	\mathcal{L}_{d} = \log \left(1 - D(\hat{y})\right) + \log D(y).
\end{equation}
We adopt Cosine Annealing Scheme~\cite{loshchilov2016sgdr} and Adam optimizer~\cite{kingma2014adam} in training. The number of iterations is 300K and the initial learning rate is $10^{-4}$. The batch size is 8 for human faces and 16 for other categories. We train our models using two Nvidia V100 GPUs.

\section{Qualitative Results}

\subsection{Super-Resolution}
\label{apx:quali}
\begin{figure*}[!t]
	\begin{center}
		\includegraphics[width=0.99\textwidth]{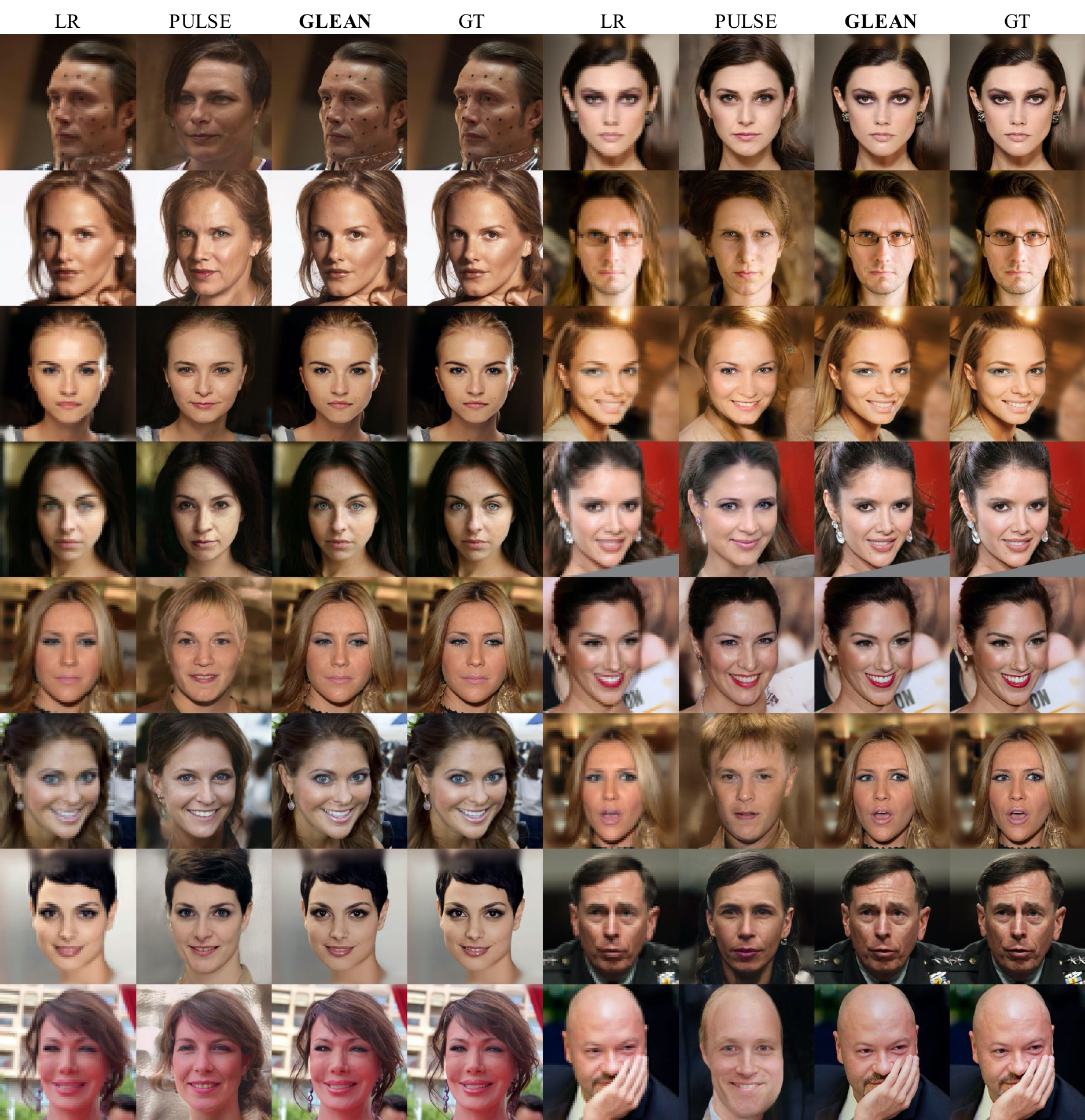}
		\caption{\textbf{Comparison to PULSE on randomly-selected examples from CelebA-HQ~\cite{karras2018progressive}.} By optimizing only the latent vectors, the outputs of PULSE~\cite{menon2020pulse} differ significantly from the ground-truths. With our novel designs, GLEAN produces outputs highly similar to the ground-truths.}
		\label{fig:pulse}
	\end{center}
\end{figure*}

\noindent\textbf{Randomly-Selected Examples.}
In Fig.~\ref{fig:pulse}, we show the results of randomly-selected examples from CelebA-HQ~\cite{karras2018progressive}. By optimizing only the latent codes, PULSE~\cite{menon2020pulse} produces outputs with low-fidelity. In contrast, guided by the encoder features and our generative latent bank, GLEAN achieves remarkable quality and fidelity, demonstrating the effectiveness of our designs.

\begin{figure*}[!t]
	\begin{center}
		\includegraphics[width=0.99\textwidth]{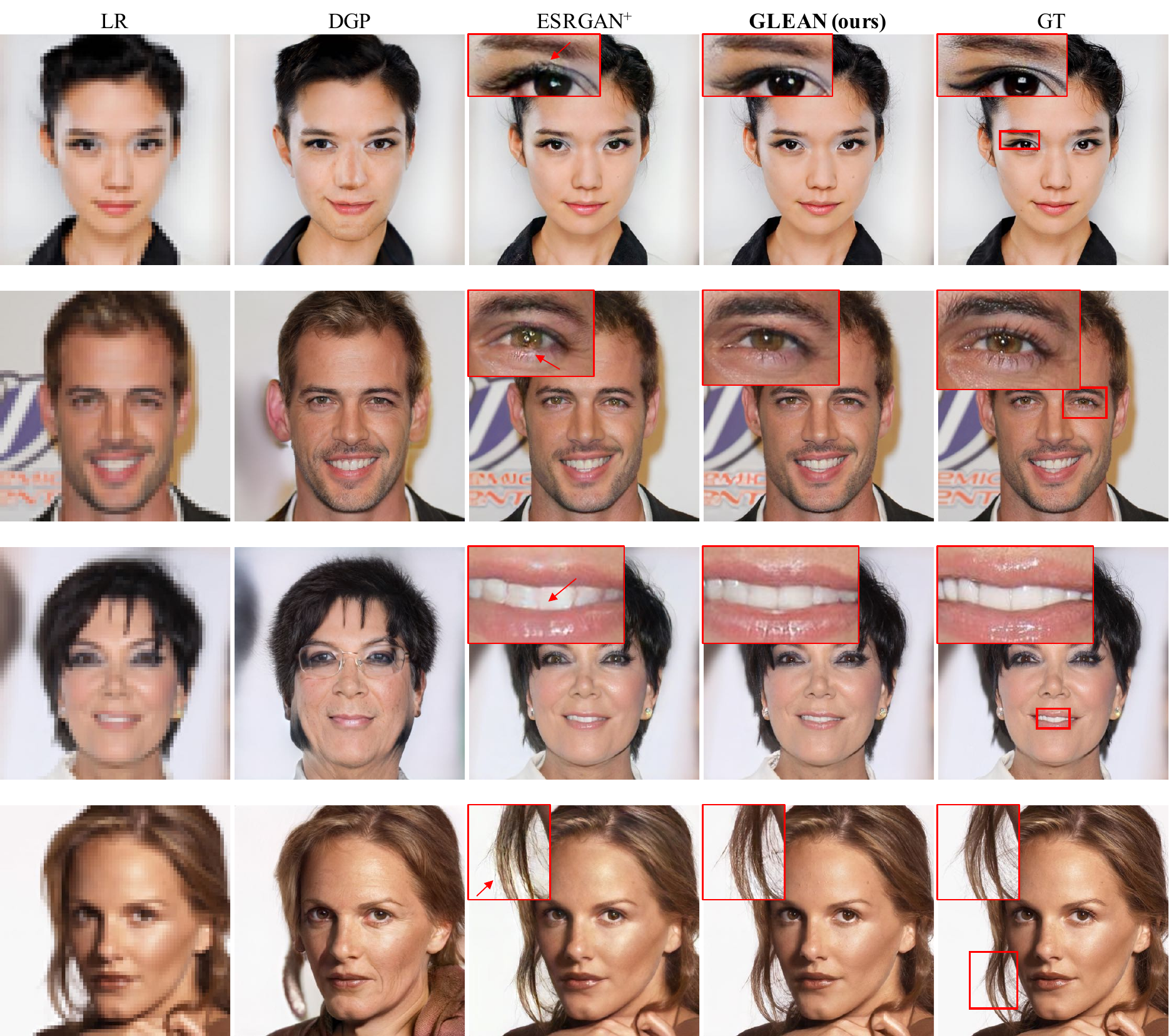}
		\caption{\textbf{Comparison with DGP and ESRGAN$^+$.} The outputs of DGP show noticeable identity differences to the ground-truths. ESRGAN$^+$ shows unpleasant artifacts for the fine details. \textbf{(Zoom-in for best view)}}
		\label{fig:face_comp}
		\vspace{-0.3cm}
	\end{center}
\end{figure*}

\begin{figure*}[!t]
	\begin{center}
		\includegraphics[width=0.99\textwidth]{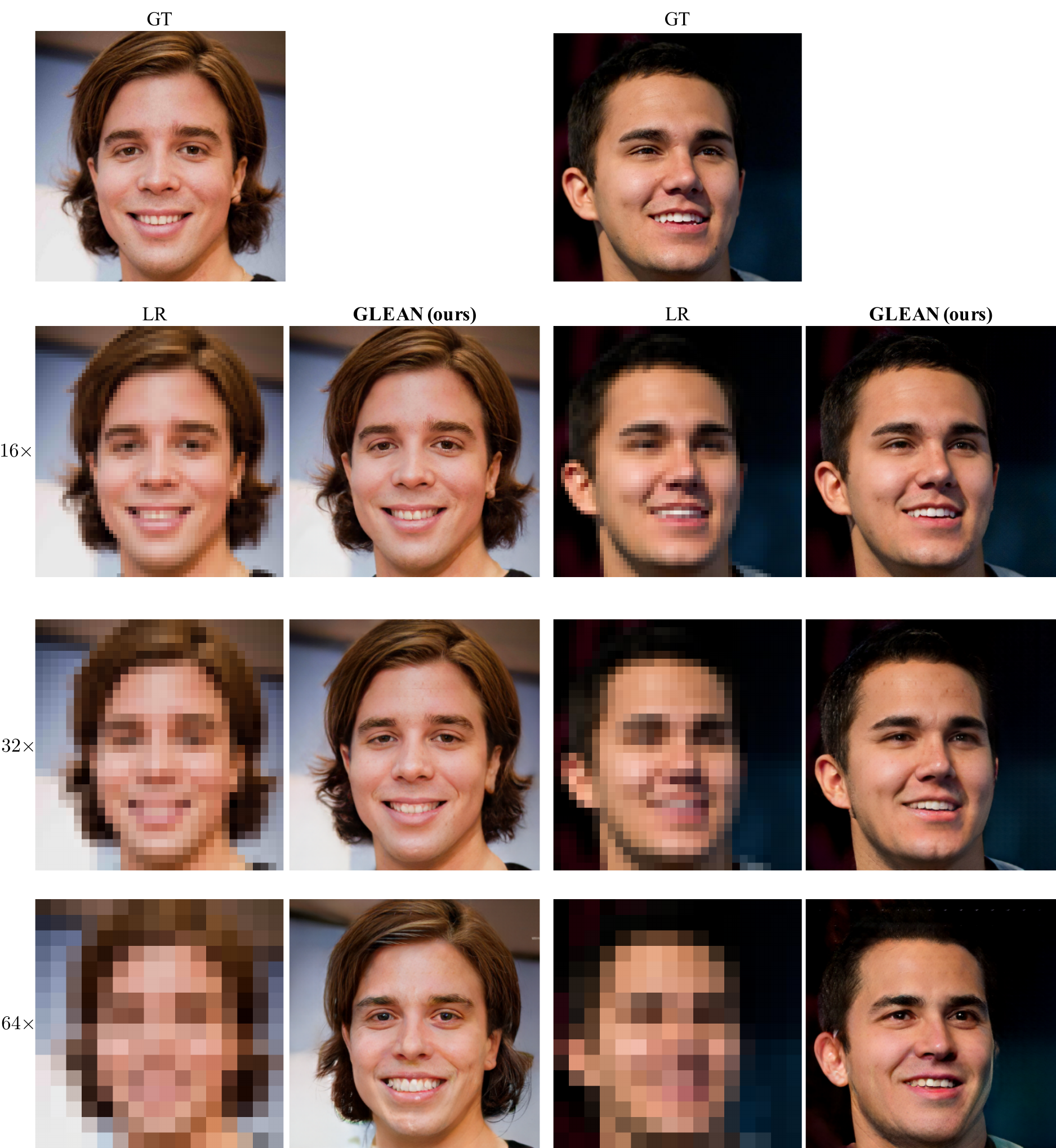}
		\caption{\textbf{\textbf{Performance of GLEAN on 16${\times}$, 32${\times}$, and 64${\times}$ SR.}} GLEAN is able to synthesize images well resembling the ground-truths for up to $64{\times}$ upsampling.}
		\label{fig:face}
		\vspace{-0.3cm}
	\end{center}
\end{figure*}

\begin{figure*}[!t]
	\begin{center}
		\includegraphics[width=0.99\textwidth]{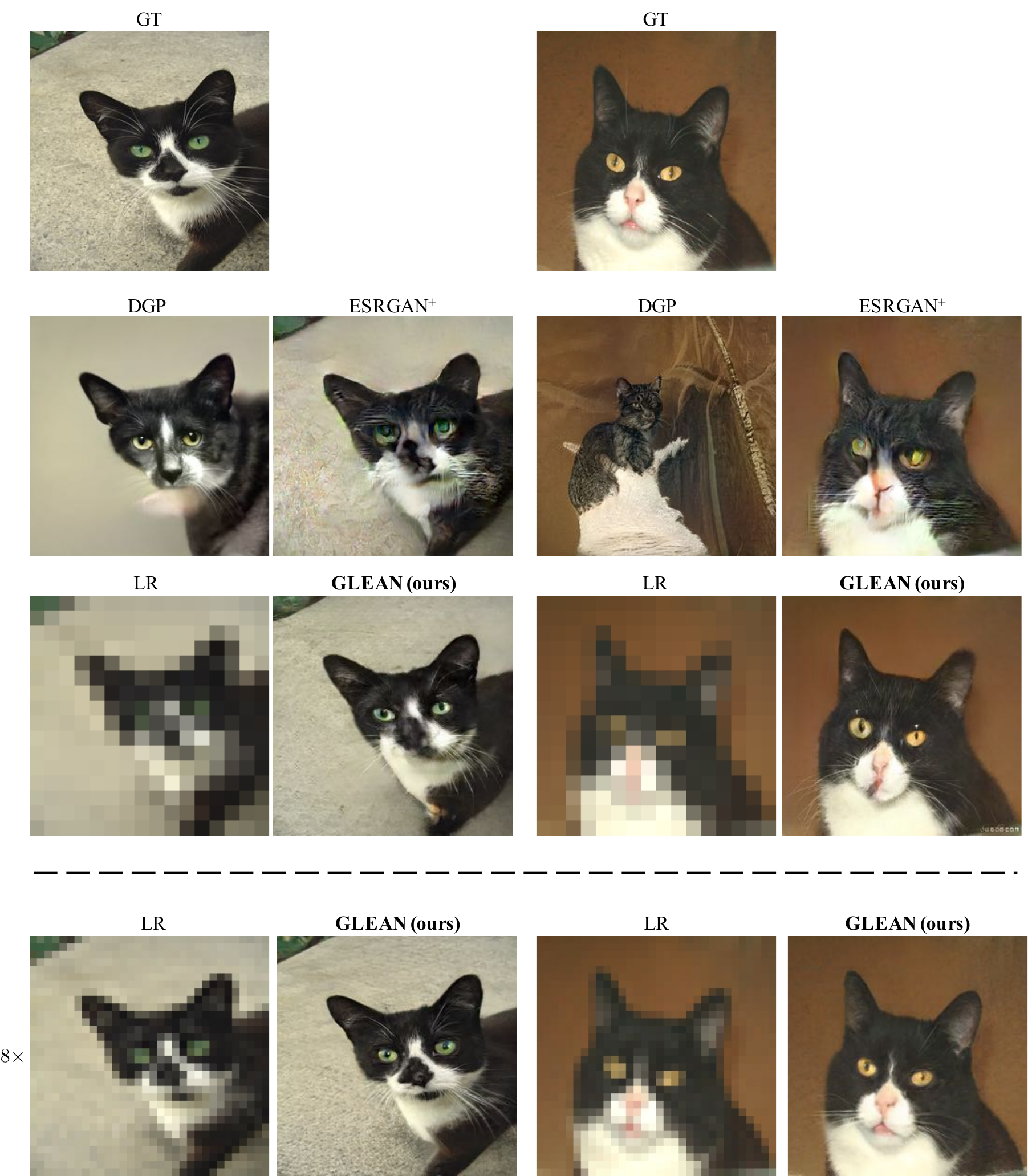}
		\caption{\textbf{(Top) Comparison with DGP and ESRGAN$^+$ on \textit{Cats}.} DGP produces outputs with low fidelity; ESRGAN$^+$ fails to synthesize realistic textures. \textbf{(Bottom) Performance of GLEAN on 8${\times}$ SR.} GLEAN produces realistic outputs that are highly similar to the ground-truths. \textbf{(Zoom-in for best view)}}
		\label{fig:cat}
	\end{center}
\end{figure*}

\begin{figure*}[!t]
	\begin{center}
		\includegraphics[width=0.99\textwidth]{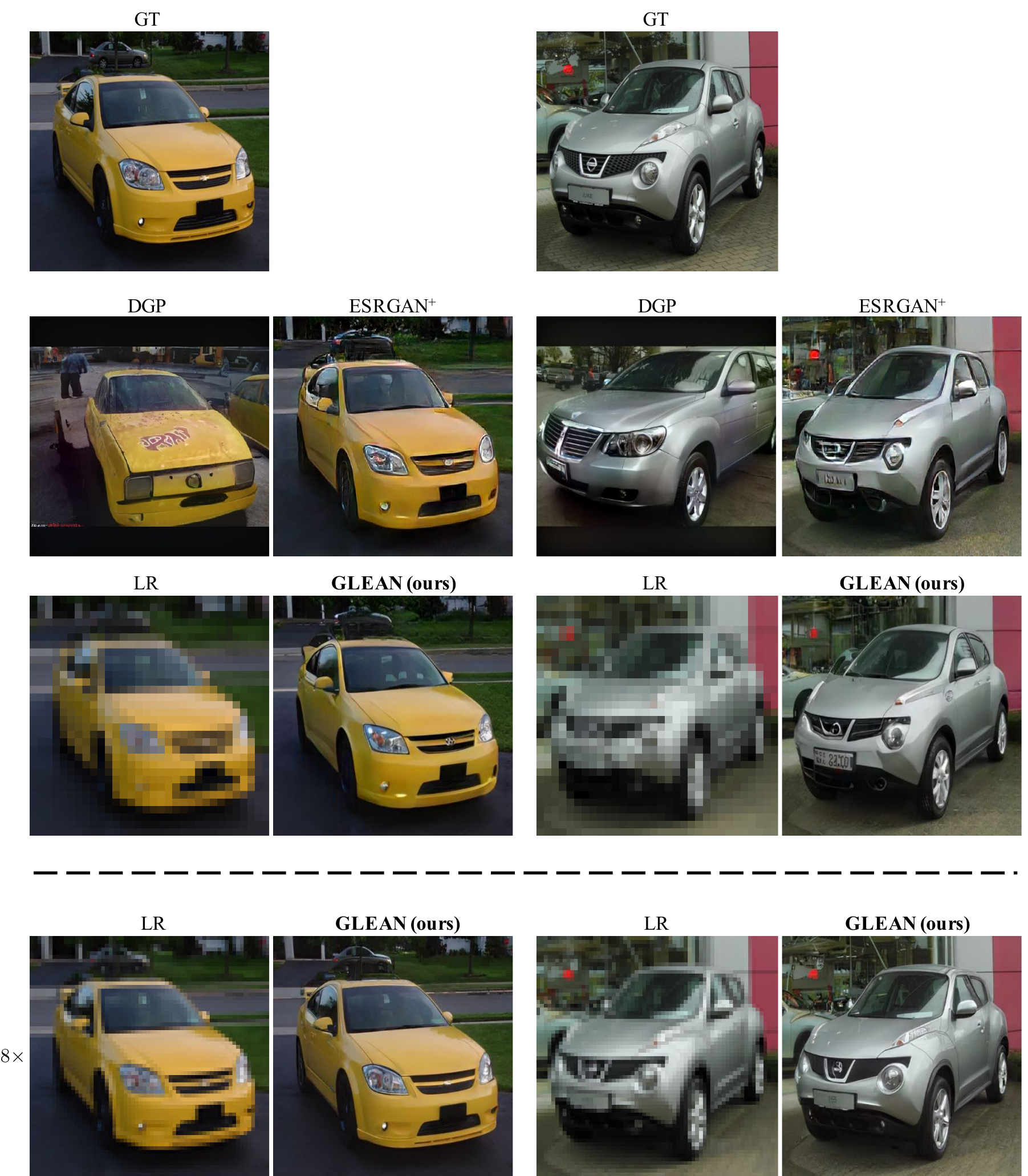}
		\caption{\textbf{(Top) Comparison with DGP and ESRGAN$^+$ on \textit{Cars}.} DGP produces outputs with low fidelity; ESRGAN$^+$ fails to synthesize realistic textures. \textbf{(Bottom) Performance of GLEAN on 8${\times}$ SR.} GLEAN produces realistic outputs that are highly similar to the ground-truths. \textbf{(Zoom-in for best view)}}
		\label{fig:car}
	\end{center}
\end{figure*}

\begin{figure*}[!t]
	\begin{center}
		\includegraphics[width=0.99\textwidth]{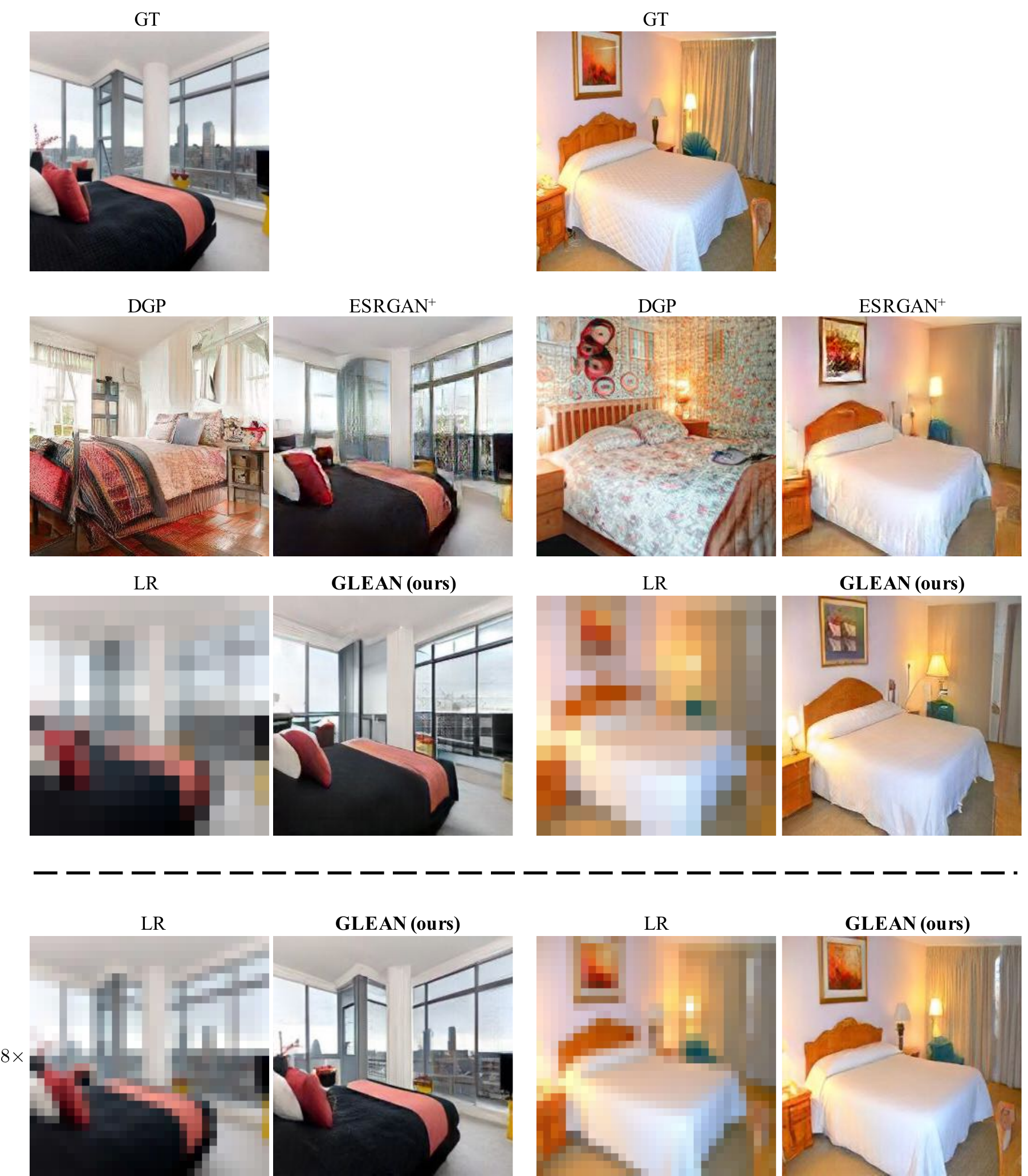}
		\caption{\textbf{(Top) Comparison with DGP and ESRGAN$^+$ on \textit{Bedrooms}.} DGP produces outputs with low fidelity; ESRGAN$^+$ fails to synthesize realistic textures. \textbf{(Bottom) Performance of GLEAN on 8${\times}$ SR.} GLEAN produces realistic outputs that are highly similar to the ground-truths. \textbf{(Zoom-in for best view)}}
		\label{fig:bedroom}
	\end{center}
\end{figure*}

\begin{figure*}[!t]
	\begin{center}
		\includegraphics[width=0.99\textwidth]{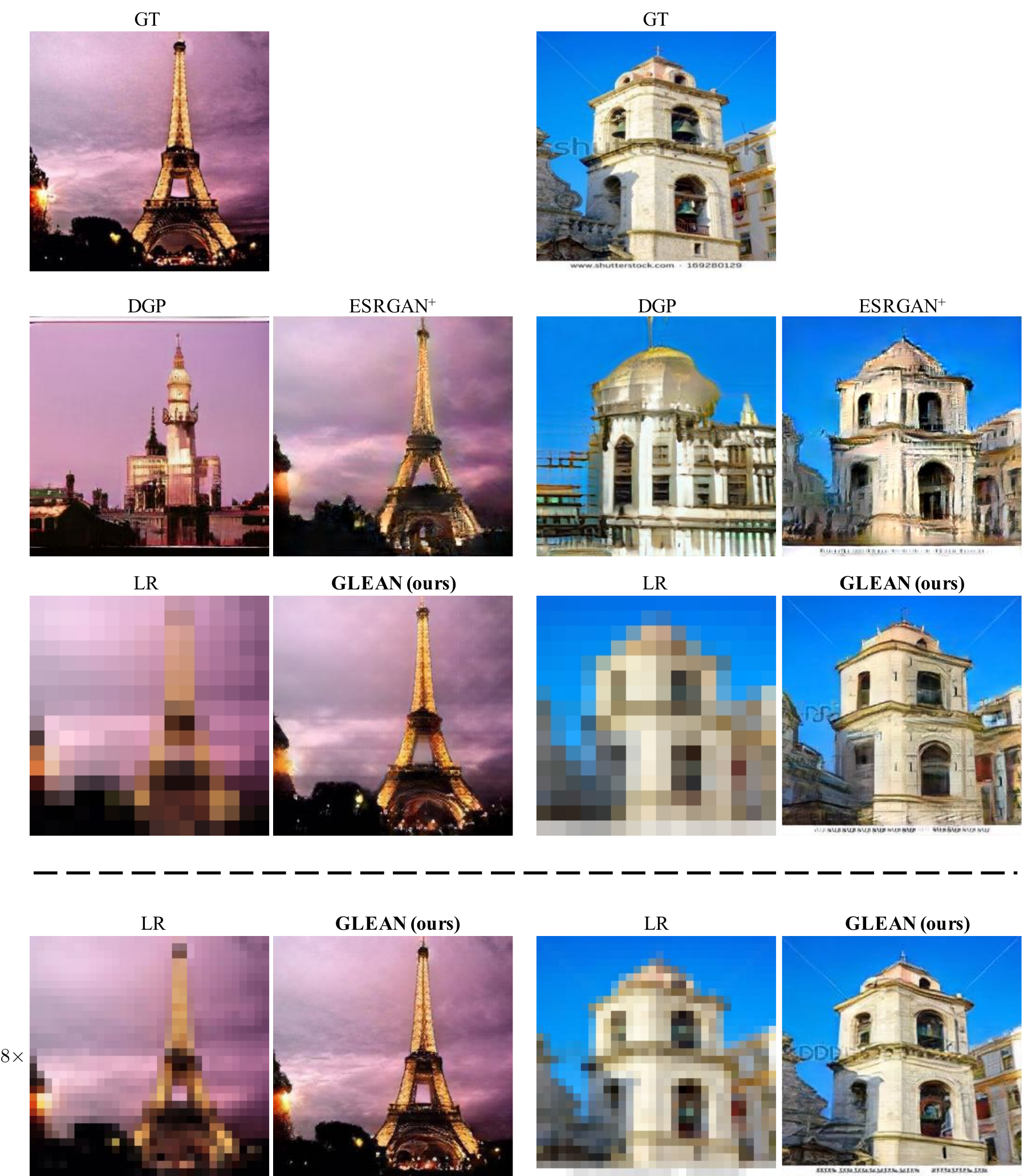}
		\caption{\textbf{(Top) Comparison with DGP and ESRGAN$^+$ on \textit{Towers}.} DGP produces outputs with low fidelity; ESRGAN$^+$ fails to synthesize realistic textures. \textbf{(Bottom) Performance of GLEAN on 8${\times}$ SR.} GLEAN produces realistic outputs that are highly similar to the ground-truths. \textbf{(Zoom-in for best view)}}
		\label{fig:tower}
	\end{center}
\end{figure*}

\noindent\textbf{Scale Factors and Categories.}
GLEAN is extensible to various scale factors (from $8{\times}$ to $64{\times}$) and categories (\eg~faces, cats, cars, bedrooms, towers). From Fig.~\ref{fig:face_comp} to Fig.~\ref{fig:tower}, we see that GLEAN outperforms DGP and ESRGAN$^+$ in both fidelity and quality. It is noteworthy that the performance of DGP and ESRGAN$^+$ are less promising on categories other than human faces.

\subsection{Image Retouching}
\label{apx:retouch}
In interactive image retouching, users can manually edit the images based on their preference. For instance, users can change the facial expression of an object and perform geometric transformations for enlarging eyes. However, a perfect output requires tedious and precise retouching. As a result, artifacts are common in the outputs from amateur retouching.

GLEAN allows the possibility of performing realistic refinement of imperfect retouching. More specifically, given a retouched image, we can first downsample the image to a smaller resolution, where the artifacts vanished. We can then upsample it back to the original resolution. With GLEAN as a powerful super-resolver, we can obtain an output with unnatural artifacts suppressed.

\begin{figure*}[!t]
	\begin{center}
		\includegraphics[width=0.95\textwidth]{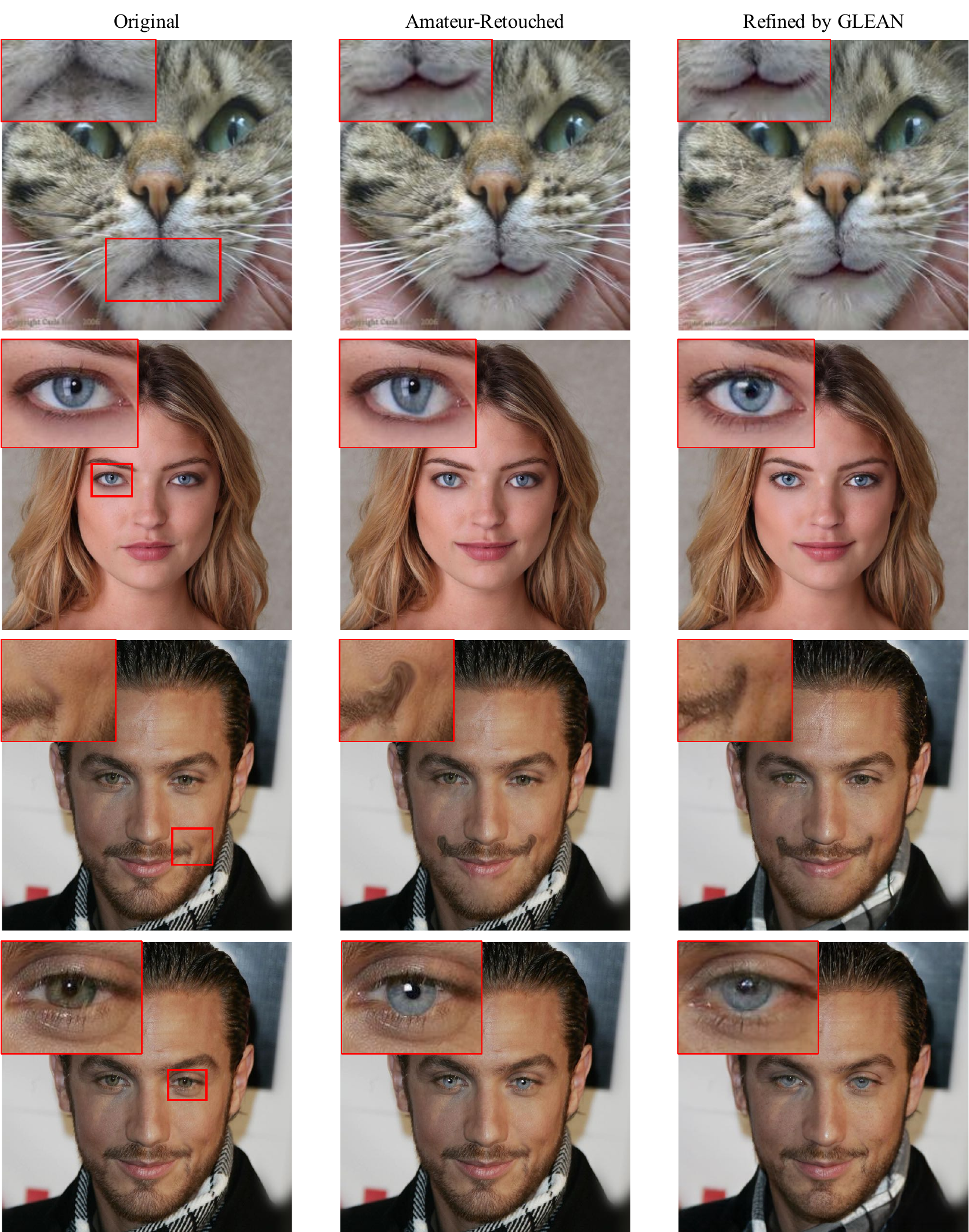}
		\caption{\textbf{Results on image retouching.} GLEAN can be used to correct unpleasant artifacts introduced by amateur retouching. \textbf{(Zoom-in for best view)}}
		\label{fig:apx_retouch}
	\end{center}
\end{figure*}

As shown in Fig.~\ref{fig:apx_retouch}, GLEAN is able to correct the unnatural artifacts introduced by amateur retouching while being similar to the retouched images, realistic, and coherent with the unaltered regions. In addition, since GLEAN requires only a single forward pass, it can be used in interactive image editing software to allow a more flexible retouching.


\end{document}